%% file: genAttention.tex
\documentclass{article}

\usepackage{times}
\usepackage{graphicx} % more modern
\usepackage{subfigure} 
\usepackage{balance}
\graphicspath{{figures/}}
\usepackage{makecell}
\usepackage{wrapfig}
\usepackage[percent]{overpic}

\usepackage{natbib}
\include{defns2}
\usepackage[export]{adjustbox}

\usepackage{enumitem}
\setlist{nosep} % or \setlist{noitemsep} to leave space around whole list
\usepackage{lipsum}
\usepackage{todonotes}

\usepackage{algorithm}
\usepackage{algorithmic}
\usepackage{hyperref}

\usepackage[accepted]{icml2016}

\icmltitlerunning{One-shot Generalization in Deep Generative Models}

\begin{document} 

\twocolumn[
%\icmltitle{Better Generative Models Using Feedback and Attention}
%\icmltitle{One Shot Generalization in Deep Generative Models Using Feedback and Attention}
\icmltitle{One-Shot Generalization in Deep Generative Models}

\icmlauthor{Danilo J. Rezende*}{danilor@google.com}
\icmlauthor{Shakir Mohamed*}{shakir@google.com}
\icmlauthor{Ivo Danihelka}{danihelka@google.com}
\icmlauthor{Karol Gregor}{karolg@google.com}
\icmlauthor{Daan Wierstra}{wierstra@google.com}
\icmladdress{Google DeepMind, London}

\icmlkeywords{deep generative models, variational inference, one-shot generalization}

%\vskip 0.3in
]

\begin{abstract} 
Humans have an impressive ability to reason about new concepts and experiences from just a single example. In particular, humans have an ability for one-shot generalization: an ability to encounter a new concept, understand its structure, and then be able to generate compelling alternative variations of the concept.
We develop machine learning systems with this important capacity by developing new deep generative models, models that combine the representational power of deep learning with the inferential power of Bayesian reasoning. We develop a class of sequential generative models that are built on the principles of feedback and attention. These two characteristics lead to generative models that are among the state-of-the art in density estimation and image generation. We demonstrate the one-shot generalization ability of our models using three tasks: unconditional sampling, generating new exemplars of a given concept, and generating new exemplars of a family of concepts. In all cases our models are able to generate compelling and diverse samples---having seen new examples just once---providing an important class of general-purpose models for one-shot machine learning. 
\vspace{-5mm}
\end{abstract} 

\input{intro}
\input{attention}
\input{models}

\input{results}
\input{oneshot}
\input{discussion}

%\balance
\input{acknowledgements}
\bibliography{refs}
\bibliographystyle{icml2016}

\nobalance
\appendix
\clearpage
\input{appdxSamples}
\input{appdxAttention}

\end{document}

%% file: defns2.tex
\usepackage{amsmath,amsthm,amssymb}
\newcommand\cut[1]{}

\newcommand{\squishlist}{
   \begin{list}{$\bullet$}
    { \setlength{\itemsep}{0pt}      \setlength{\parsep}{3pt}
      \setlength{\topsep}{3pt}       \setlength{\partopsep}{0pt}
      \setlength{\leftmargin}{1.5em} \setlength{\labelwidth}{1em}
      \setlength{\labelsep}{0.5em} } }

\newcommand{\squishlisttwo}{
   \begin{list}{$\bullet$}
    { \setlength{\itemsep}{0pt}    \setlength{\parsep}{0pt}
      \setlength{\topsep}{0pt}     \setlength{\partopsep}{0pt}
      \setlength{\leftmargin}{2em} \setlength{\labelwidth}{1.5em}
      \setlength{\labelsep}{0.5em} } }

\newcommand{\squishend}{
    \end{list}  }

\newcommand{\KL}{\mbox{KL}}

\newcommand{\myvec}[1]{\mathbf{#1}}
\newcommand{\myvecsym}[1]{\boldsymbol{#1}}

\newcommand{\vzero}{\myvecsym{0}}

\newcommand{\vmu}{\myvecsym{\mu}}

\newcommand{\vlambda}{\myvecsym{\lambda}}

\newcommand{\vsigma}{\myvecsym{\sigma}}

\newcommand{\vc}{\myvec{c}}

\newcommand{\vh}{\myvec{h}}

\newcommand{\vr}{\myvec{r}}
\newcommand{\vs}{\myvec{s}}

\newcommand{\vu}{\myvec{u}}
\newcommand{\vv}{\myvec{v}}

\newcommand{\vx}{\myvec{x}}

\newcommand{\vy}{\myvec{y}}

\newcommand{\vz}{\myvec{z}}

\newcommand{\vB}{\myvec{B}}

\newcommand{\vI}{\myvec{I}}

\newcommand{\vU}{\myvec{U}}

\newcommand{\be}{\begin{equation}}
\newcommand{\ee}{\end{equation}}
\newcommand{\bea}{\begin{eqnarray}}
\newcommand{\eea}{\end{eqnarray}}
\newcommand{\beaa}{\begin{eqnarray*}}
\newcommand{\eeaa}{\end{eqnarray*}}

\DeclareMathAlphabet{\mathpzc}{OT1}{pzc}{m}{n}

%% file: intro.tex
\section{Introduction}
\vspace{-3mm}
\begin{wrapfigure}{l}{0.18\textwidth}
\centering
\vspace{-5mm}
\hspace{-15mm}
    \includegraphics[width=0.18\textwidth]{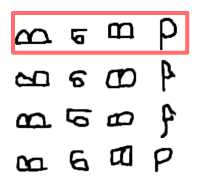}
\hspace{-18mm}
    \vspace{-5mm}
  \caption{Given the first row, our model generates new exemplars. 
  %Extract from figure \ref{fig:oneshot-generation-weak}.
  }
  \label{fig:introGen}
  \vspace{-7mm}
 %\hspace{-10mm}
\end{wrapfigure}
Consider the images in the red box in figure \ref{fig:introGen}. We see each of these new concepts just once, understand their structure, and are then able to imagine and generate compelling alternative variations of each concept, similar to those drawn in the rows beneath the red box. This is an ability that humans have for \textit{one-shot generalization}: an ability to generalize to new concepts given just one or a few examples. In this paper, we develop new models that possess this capacity for one-shot generalization---models that allow for one-shot reasoning from the data streams we are likely to encounter in practice, that use only limited forms of domain-specific knowledge, and  that can be applied to diverse sets of problems.
%This is a powerful ability that is demonstrated by few machine learning methods
%We develop new models that have this pro that allows us to exploit one-shot reasoning that allows us to   
%This capacity is one-shot reasoning is exhibited in many different ways. The most commonly explored type in machine learning is that of one-shot classification and recognition, but another important, though less-explored type, is that of \textit{one-shot generalization}. 
%To exploit these types of one-shot reasoning when learning from the data streams we are likely to encounter in practice, we require approaches that are general-purpose, using only limited forms of domain-specific assumptions, allowing them to be applied to the widest range of problems possible. 
%We develop such models in this paper. 

There are two notable approaches that incorporate one-shot generalization. \citet{salakhutdinov2013learning} developed a probabilistic model that combines a deep Boltzmann machine with a hierarchical Dirichlet process to learn hierarchies of concept categories as well as provide a powerful generative model. Recently, \citet{lake2015human}  presented a compelling demonstration of the ability of probabilistic models to perform one-shot generalization, using Bayesian program learning, which is able to learn a hierarchical, non-parametric generative model of handwritten characters. Their approach incorporates specific knowledge of how strokes are formed and the ways in which they are combined to produce characters of different types, exploiting similar strategies used by humans. 
\citet{lake2015human} see the capacity for one-shot generalization demonstrated by Bayesian programming learning `as a challenge for neural models'. By combining the representational power of deep neural networks embedded within hierarchical latent variable models, with the inferential power of approximate Bayesian reasoning, we show that this is a challenge that can be overcome.
%To address this questions, we develop a class of generative models that incorporates deep neural networks with hierarchical latent variable models and Bayesian inference. 
The resulting deep generative models are general-purpose image models that are accurate and scalable, among the state-of-the-art, and possess the important capacity for one-shot generalization.

Deep generative models are a rich class of models for density estimation that specify a generative process for observed data using a hierarchy of latent variables. Models that are directed graphical models have risen in popularity and include discrete latent variable models such as sigmoid belief networks and deep auto-regressive networks \citep{saul1996mean, gregor2013deep}, or continuous latent variable models such as non-linear Gaussian belief networks and deep latent Gaussian models \citep{rezende2014stochastic, kingma2014stochastic}. These models use deep networks in the specification of their conditional probability distributions to allow rich non-linear structure to be learned. Such models have been shown to have a number of desirable properties: inference of the latent variables allows us to provide a causal explanation for the data that can be used to explore its underlying factors of variation and for exploratory analysis; analogical reasoning between two related concepts, e.g., styles and identities of images, is naturally possible; any missing data can be imputed by treating them as additional latent variables, capturing the the full range of correlation between missing entries under any missingness pattern; these models embody minimum description length principles and can be used for compression; these models can be used to learn environment-simulators enabling a wide range of approaches for simulation-based planning.

Two principles are central to our approach: feedback and attention. These principles allow the models we develop to reflect the principles of \textit{analysis-by-synthesis}, in which the analysis of observed information is continually integrated with constructed interpretations of it \citep{yuille2006vision, erdogan2015, nair2008analysis}. Analysis is realized by attentional mechanisms that allow us to selectively process and route information from the observed data into the model. Interpretations of the data are then obtained by sets of latent variables that are inferred sequentially to evaluate the probability of the data. 
The aim of such a construction is to introduce internal feedback into the model that allows for a `thinking time' during which information can be extracted from each data point more effectively, leading to improved inference, generation and generalization.
We shall refer to such models as \textit{sequential generative models}.
% are able to better incorporate information from each data point, leading to improved inference, generation and generalization. 
Models such as DRAW \citep{gregor2015draw}, composited variational auto-encoders \citep{huang2015efficient} and AIR \citep{eslami2016attend} are existing models in this class, and we will develop a general class of sequential generative models that incorporates these and other latent variable models and variational auto-encoders.

Our contributions are:
\vspace{-2mm}
\begin{itemize}[leftmargin=*]
\item We develop sequential generative models that provide a generalization of existing approaches, allowing for sequential generation and inference, multi-modal posterior approximations, and a rich new class of deep generative models.
%\item  We develop and compare various attentional mechanisms that provide highly flexible and scalable attention.   
\item We demonstrate the clear improvement that the combination of attentional mechanisms in more powerful models and inference has in advancing the state-of-the-art in deep generative models. 
\item Importantly, we show that our generative models have the ability to perform one-shot generalization. We explore three generalization tasks and show that our models can imagine and generate compelling alternative variations of images after having seen them just once.
\end{itemize}

%% file: attention.tex
\vspace{-2mm}
\section{Varieties of Attention}
\label{sect:attention}
Attending to parts of a scene, ignoring others, analyzing the parts that we focus on, and sequentially building up an interpretation and understanding of a scene: these are natural parts of human cognition. This is so successful a strategy for reasoning that it is now also an important part of many machine learning systems. This repeated process of attention and interpretation, analysis and synthesis, is an important component of the generative models we develop.

In its most general form, any mechanism that allows us to selectively route information from one part of our model to another can be regarded as an attentional mechanism. 
%Sparse selection methods, such as selecting a patch of an image or input-dimension subsampling, and methods that provide a coordinate transformations into a canonical representation, such as rotations and shifts for images or band-pass filters, are examples of attentional mechanisms. These mechanisms can be deterministic or stochastic, differentiable or non-differentiable. Furthermore, attentional mechanisms go beyond providing a generic dimensionality reduction components to the model. 
Attention allows for a wide range of invariances to be incorporated, with few additional parameters and low computational cost. 
Attention has been most widely used for classification tasks, having been shown to improve both scalability and generalization \cite{larochelle2010learning, chikkerur2010and, xu2015show, jaderberg2015spatial, mnih2014recurrent, ba2015learning}. The attention used in discriminative tasks is a `reading' attention that transforms an image into a representation in a canonical coordinate space (that is typically lower dimensional), with the parameters controlling the attention learned by gradient descent.
%, not on the image itself, but on a transformation of the image (into a canonical coordinate space that is typically lower dimensional), with the parameters controlling the attention being learned by gradient descent.
Attention in unsupervised learning is much more recent \citep{tang2014learning, gregor2015draw}. In latent variable models, we have two processes---inference and generation---that can both use attention, though in slightly different ways. The generative process makes use of a \textit{writing or generative attention}, which implements a selective updating of the output variables, e.g., updating only a small part of the generated image. The inference process makes use of \textit{reading attention}, like that used in classification. Although conceptually different, both these forms of attention can be implemented with the same computational tools.
We focus on image modelling and make use of spatial attention. 
Two other types of attention, randomized and error-based, are discussed in appendix \ref{appdx:attention}.
\\  \\
\textbf{\textit{Spatially-transformed attention}}. Rather than selecting a patch of an image (taking glimpses) as other methods do, a more powerful approach is to use a mechanism that provides invariance to shape and size of objects in the images (general affine transformations).  \citet{tang2014learning} take such an approach and use 2D similarity transforms to provide basic affine invariance. Spatial transformers \citep{jaderberg2015spatial} are a more general method for providing such invariance, and is our preferred attentional mechanism. Spatial transformers (ST) process an input image $\vx$ using parameters $\vlambda$ to generate an output:
%\begin{align}
$$ \textrm{ST}(\vx, \vlambda) = \left[ \kappa_h(\vlambda) \otimes \kappa_w(\vlambda) \right]   \ast  \vx, \nonumber
$$
%\end{align}
where $\kappa_h$ and $\kappa_w$ are 1-dimensional kernels, $\otimes$ indicates the tensor outer-product of the two kernels and $\ast$ indicates a convolution. 
\citet{huang2015efficient} develop occlusion-aware generative models that make use of spatial transformers in this way. When used for reading attention, spatial transformers allow the model to observe the input image in a canonical form, providing the desired invariance. When used for writing attention, it allows the generative model to independently handle position, scale and rotation of parts of the generated image, as well as their content. An direct extension is to use multiple attention windows simultaneously (see appendix).

%% file: models.tex
\vspace{-2mm}
\section{Iterative and Attentive Generative Models}
\vspace{-2mm}
\subsection{Latent Variable Models and Variational Inference}
Generative models  with latent variables describe the probabilistic process by which an observed data point can be generated. The simplest formulations such as PCA and factor analysis use Gaussian latent variables $\vz$ that are combined linearly to generate Gaussian distributed data points $\vx$. In more complex models, the probabilistic description consists of a hierarchy of $L$ layers of latent variables, where each layer depends on the layer above in a non-linear way \cite{rezende2014stochastic}. For deep generative models, we specify this non-linear dependency using deep neural networks. To compute the marginal probability of the data, we must integrate over any unobserved variables:\
\vspace{-2mm}
\begin{equation}
p(\vx) = \int p_\theta(\vx | \vz) p(\vz) d\vz \label{eq:marglik}
\vspace{-3mm}
\end{equation}
In deep latent Gaussian models, the prior distribution $p(\vz)$ is a Gaussian distribution and the likelihood function $p_\theta(\vx | \vz)$ is any distribution that is appropriate for the observed data, such as a Gaussian, Bernoulli, categorical or other distribution, and that is dependent in a non-linear way on the latent variables. For most models, the marginal likelihood \eqref{eq:marglik} is intractable and we must instead approximate it. One popular approximation technique is based on variational inference \citep{jordan1999introduction}, which transforms the difficult integration into an optimization problem that is typically more scalable and easier to solve.
Using variational inference we can approximate the marginal likelihood by a lower bound, which is the objective function we use for optimization:
\begin{eqnarray}
\mathcal{F} = \mathbb{E}_{q(\vz|\vx)}[\log p_\theta(\vx | \vz)] - \KL[q_\phi(\vz | \vx) \| p(\vz)] \label{eq:FE}
\end{eqnarray}
The objective function \eqref{eq:FE} is the negative free energy, which allows us to trade-off the reconstruction ability of the model (first term) against the complexity of the posterior  distribution (second term). Variational inference approximates the true posterior distribution by a known family of approximating posteriors $q_\phi(\vz | \vx)$ with variational parameters $\phi$. Learning now involves optimization of the variational parameters $\phi$ and model parameters $\theta$. 

Instead of optimization by the variational EM algorithm, we take an amortized inference approach and represent the distribution $q(\vz | \vx)$ as a recognition or inference model, which we also parameterize using a deep neural network. Inference models amortize the cost of posterior inference and makes it more efficient by allowing for generalization across the inference computations using a set of global variational parameters $\phi$. In this framework, we can think of the generative model as a decoder of the latent variables, and the inference model as its inverse, an encoder of the observed data into the latent description. As a result, this specific combination of deep latent variable model (typically latent Gaussian) with variational inference that is implemented using an inference model is referred to as a variational auto-encoder (VAE). VAEs allow for a single computational graph to be constructed and straightforward gradient computations: when the latent variables are continuous, gradient estimators based on pathwise derivative estimators are used \citep{rezende2014stochastic, kingma2014stochastic, burda2015importance} and when they are discrete, score function estimators are used \citep{mnih2014neural, ranganath2013black, mansimov2015generating}.
\vspace{-2mm}
\subsection{Sequential Generative Models}
The generative models as we have described them thus far can be characterized as single-step models, since they are models of i.i.d data that evaluate their likelihood functions by transforming the latent variables using a non-linear, feed-forward transformation. A \textit{sequential generative model} is a natural extension of the latent variable models used in VAEs. Instead of generating the $K$ latent variables of the model in one step, these models sequentially generate $T$ groups of $k$ latent variables ($K= kT$), i.e. using $T$ computational steps to allow later groups of latent variables to depend on previously generated latent variables in a non-linear way.
%They combine both stochastic and deterministic computations to form a multi-step generative process that uses recursive transformations of the latent variables, i.e. uses an internal state-space model.
\begin{figure*}[t!]
 \centering
\subfigure[Unconditional generative model.]{
\label{fig:uncondModel}
\includegraphics[height=4cm]{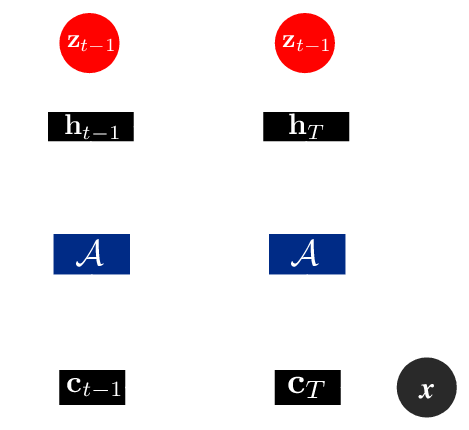}
\hspace{0.06cm}
\includegraphics[height=4cm]{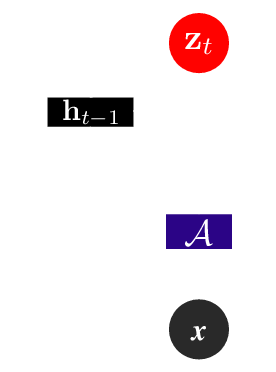}
}
\hspace{0.6cm}
\subfigure[One-step of the conditional generative model.]{
\includegraphics[height=4cm]{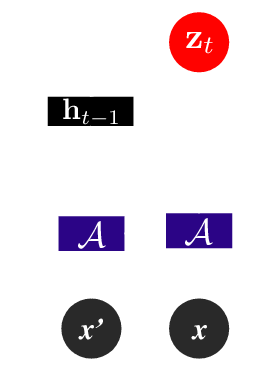}
\label{fig:condModel}
\hspace{0.1cm}
\includegraphics[height=4cm]{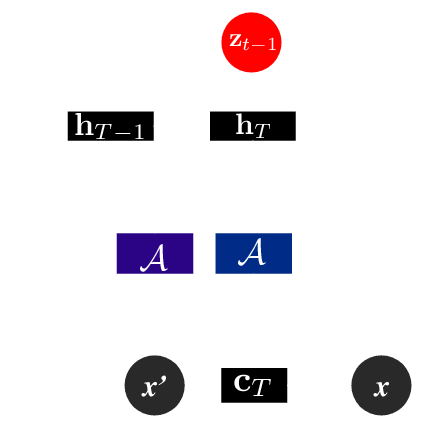}
}
\vspace{-4mm}
\caption{Stochastic computational graph showing conditional probabilities and computational steps for sequential generative models. $\mathcal{A}$ represents an attentional mechanism that uses function $f_w$ for writings and function $f_r$ for reading.}
%=======
%\caption{Stochastic computational graph showing conditional probabilities and computational steps for sequential generative models. \mbox{$\mathcal{A}$} indicates where an attention mechanism is used. }
%>>>>>>> 2016_ICML_GenAttention
\vspace{-5mm}
\end{figure*}
\vspace{-2mm}
\subsubsection{Generative Model}
In their most general form, sequential generative models describe the observed data over $T$ time steps using a set of latent variables $\vz_t$ at each step. The generative model is shown in the stochastic computational graph of figure \ref{fig:uncondModel}, and described by:
\vspace{-2mm}
\begin{eqnarray}
\textrm{Latent variables} & \vz_t  \sim & \mathcal{N}(\vz_t | \vzero, \vI) \,\,\, t = 1, \dots, T \label{eq:mod_latent} \\
\textrm{Context} & \vv_t  = & f_v(\vh_{t-1}, \vx'; \theta_v) \label{eq:mod_context}\\
\textrm{Hidden state} & \vh_t  = & f_h(\vh_{t-1}, \vz_t, \vv_t; \theta_h)  \label{eq:mod_hid} \\
\textrm{Hidden Canvas} & \vc_t  = & f_c(\vc_{t-1}, \vh_t;\theta_c) \label{eq:mod_canvas} \\
\textrm{Observation} & \vx  \sim &p(\vx | f_o(\vc_T; \theta_o) ) \label{eq:mod_obs}
\vspace{-2mm}
\end{eqnarray}
Each step generates an independent set of $K$-dimensional latent variables $\vz_t$ (equation \eqref{eq:mod_latent}). 
If we wish to condition the model on an external context or piece of side-information $\vx'$, then a deterministic function $f_v$ (equation \eqref{eq:mod_context}) is used to read the context-images using an attentional mechanism. 
A deterministic transition function $f_h$ introduces the sequential dependency between each of the latent variables, incorporating the context if it exists (equation \eqref{eq:mod_hid}). This allows any transition mechanism to be used and our transition is specified as a long short-term memory network (LSTM, \citet{hochreiter1997long}. We explicitly represent the creation of a set of hidden variables $\vc_t$ that is a \textit{hidden canvas} of the model (equation \eqref{eq:mod_canvas}). The canvas function $f_c$ allows for many different transformations, and it is here where generative (writing) attention is used; we describe a number of choices for this function in section \ref{sect:canvasFn}. 
The generated image \eqref{eq:mod_obs} is sampled using an observation function $f_o(\vc; \theta_o)$ that maps the last hidden canvas $\vc_T$ to the parameters of the observation model.
%, e.g., probability of being on or off for a Bernoulli distribution. 
The set of all parameters of the generative model is $\theta = \{ \theta_h, \theta_c, \theta_o \}$.
\subsubsection{Free Energy Objective}
Given the probabilistic model \eqref{eq:mod_latent}-\eqref{eq:mod_obs} we can obtain an objective function for inference and parameter learning using variational inference. By applying the variational principle, we obtain the free energy objective:
\begin{eqnarray}
& \log p(\vx) = \log \int p(\vx | \vz_{1:T})p(\vz_{1:T}) d\vz_{1:T} \geq \mathcal{F} \nonumber \\
& \mathcal{F} = \mathbb{E}_{q(\vz_{1:T})}[\log p_\theta(\vx | \vz_{1:T} )] \nonumber \\
& - \sum_{t=1}^T \KL[q_\phi(\vz_t | \vz_{<t} \vx) \| p(\vz_t)],
\end{eqnarray}
where $\vz_{<t}$ indicates the collection of all latent variables from step 1 to $t-1$.
We can now optimize this objective function for the variational parameters $\phi$ and the model parameters $\theta$, by stochastic gradient descent using a mini-batch of data. As with other VAEs, we use a single sample of the latent variables generated from $q_{\phi}(\vz | \vx)$ when computing the Monte Carlo gradient. 
To complete our specification, we now specify the hidden-canvas functions $f_c$ and the approximate posterior distribution $q_\phi(\vz_t)$.
% we now describe specific forms of the hidden-canvas functions $f_c$ and the approximate posterior distribution $q_\phi(\vz_t)$.

\subsubsection{Hidden Canvas functions}
\label{sect:canvasFn}
The canvas transition function $f_c(\vc_{t-1}, \vh_t;\theta_c)$  \eqref{eq:mod_canvas} updates the hidden canvas by first non-linearly transforming the current hidden state of the LSTM $\vh_t$ (using a function $f_w$) and fuses the result with the existing canvas $\vc_{t-1}$. In this work we use hidden canvases that have the same size as the original images, though they could be either \textit{larger or smaller} in size and can have any number of channels (four in this paper). We consider two ways with which to update the hidden canvas:

\textit{\textbf{Additive Canvas.}} As the name implies, an additive canvas updates the canvas by simply adding a transformation of the hidden state $f_w(\vh_t;\theta_c )$ to the previous canvas state $\vc_{t-1}$. This is a simple, yet effective (see results) update rule:
\begin{equation}
f_c(\vc_{t-1}, \vh_t;\theta_c) = \vc_{t-1} + f_w(  \vh_t;\theta_c ), \label{eq_additive_canvas}
\end{equation}
\textit{\textbf{Gated Recurrent Canvas.}} The canvas function can be updated using a convolutional gated recurrent unit (CGRU) architecture \cite{kaiser2015neural}, which provides a non-linear and recursive updating mechanism for the canvas and are simplified versions of convolutional LSTMs (further details of the CGRU are given in appendix \ref{appdx:attention}). The canvas update is:
\begin{equation}
f_c(\vc_{t-1}, \vh_t;\theta_c) = \text{CGRU}( \vc_{t-1} +  f_w(  \vh_t;\theta_c ) ) \label{eq_cgru_canvas}
\end{equation}
In both cases, the function $f_w(  \vh_t;\theta_w)$ is a \textit{writing} or \textit{generative attention} function, that we implement as a spatial transformer; inputs to the spatial transformer are its affine parameters and a \mbox{$10 \times 10$} image to be transformed, both of which are provided by the LSTM output.
%the location in the hidden canvas to which it writes, both of which are provided by the LSTM output.\todo{please check here on use of ST}

The final phase of the generative process transforms the hidden canvas at the last time step $\vc_T$ into the parameters of the likelihood function using the output function  $f_o(\vc; \theta_o)$. Since we use a hidden canvas that is the same size as the original images but that have a different number of filters, we implement the output function as a \mbox{$1 \times 1$} convolution. When the hidden canvas has a different size, a convolutional network is used instead.

\subsubsection{Dependent Posterior Inference}
We use a structured posterior approximation that has an auto-regressive form, i.e. $q(\vz_t | \vz_{<t}, \vx)$. We implement this distribution as an inference network parameterized by a deep network. The specific form we use is:
\begin{eqnarray}
\textrm{Sprite } & \vr_t  =  f_r(\vx, \vh_{t-1}; \phi_r)\\
\textrm{Sample } & \!\!\!\!\vz_t \! \sim \! \mathcal{N}(\vz_t | \vmu(\vs_t, \!\vh_{t-1};\! \phi_\mu),\! \sigma(\vr_t, \!\vh_{t-1}; \phi_\sigma)\!)
\end{eqnarray}
At every step of computation, we form a low-dimensional representation $\vr_t$ of the input image using a non-linear transformation $f_r$ of the input image and the hidden state of the model.%\todo{check here for ST in inf, and posterior.} 
This function is \textit{reading} or \textit{recognition attention} using a spatial transformer, whose affine parameters are given by the LSTM output. 
 The result of reading is a sprite $\vr_t$ that is then combined with the previous state $\vh_{t-1}$ through a further non-linear function to produce the mean $\vmu_t$ and variance $\vsigma_t$ of a $K$-dimensional diagonal Gaussian distribution. We denote all the parameters of the inference model by $\phi = \{\phi_r, \phi_\mu, \phi_\sigma\}$. Although the conditional distributions $q(\vz_t | \vz_{<t})$ are Gaussian, the joint posterior posterior $p(\vz_{1:T}) = \prod_t p(\vz_t |\vz_{<t})$ is non-Gaussian and multi-modal due to the non-linearities used, enabling more accurate inference.

\subsubsection{Model Properties and Complexity}
The above sequential generative model and inference is a generalization of existing models such as DRAW \citep{gregor2015draw} , composited VAEs \citep{huang2015efficient} and AIR \citep{eslami2016attend}. This generalization has a number of differences and important properties. One of the largest deviations is the introduction of the hidden canvas into the generative model that provides an important richness to the model, since it allows a pre-image to be constructed in a hidden space before a final corrective transformation, using the function $f_o$, is used. The generative process has an important property that allows the model be sampled without feeding-back the results of the canvas $\vc_t$ to the hidden state $\vh_t$---such a connection is not needed and provides more efficiency by reducing the number of model parameters. The inference network in our framework is also similarly simplified. We do not use a separate recurrent function within the inference network (like DRAW), but instead share parameters of the LSTM from the prior---the removal of this additional recursive function has no effect on performance.

Another important difference between our framework and existing frameworks is the type of attention that is used. \citet{gregor2015draw} use a generative attention based on Gaussian convolutions parameterized by a location and scale, and \citet{tang2014learning} use 2D similarity transformations. We use a much more powerful and general attention mechanism based on spatial transformers \cite{jaderberg2015spatial, huang2015efficient}.

The overall complexity of the algorithm described matches the typical complexity of widely-used methods in deep learning. For images of size $I \times I$, the spatial transformer has a complexity that is linear in the number of pixels of the attention window. For a $J \times J$ attention window, with $J \leq I$, the spatial transformer has a complexity of $O(NTJ^2)$, for $T$ sequential steps and $N$ data points. All other components have the standard quadratic complexity in the layer size, hence for $L$ layers with average size $D$, this gives a complexity of $O(NLD^2)$. 
%\com{there should be some terms with complexity in K, size of latents as well?}

%% file: results.tex
\vspace{-3mm}
\section{Image Generation and Analysis} \label{sec.results}
\vspace{-1mm}
%We first establish the modelling power of sequential generative models. 
We first show that our models are state-of-the-art, obtaining highly competitive likelihoods, and are able to generate high-quality samples across a wide range of data sets with different characteristics. % (such as the amount of data and dimensionality).

For all our experiments, our data consists of binary images and we use use a Bernoulli likelihood to model the probability of the pixels. In all models we use 400 LSTM hidden units. We use $12 \times 12$ kernels for the spatial transformer, whether used for recognition or generative attention. The latent variable $\vz_t$ are 4-dimensional Gaussian distributions and we use a number of steps that vary from 20-80. The hidden canvas has dimensions that are the size of the images with four channels. 
%We use one sample from the approximate posterior per time-step to compute the gradients and use a batch size of 24. 
We present the main results here and any additional results in Appendix \ref{appdx:addResults}. All the models were trained for approximatively 800K iterations with mini-batches of size 24. 
%We report negative log-likelihood bounds for train and test samples in nats. 
The reported likelihood bounds for the training set are computed by averaging the last 1K iterations during training. The reported likelihood bounds for the test set were computed by averaging the bound for $24,000$ random samples (sampled with replacement) and the error bars are the standard-deviations of the mean.

\subsection{MNIST and Multi-MNIST}
\begin{figure}[t]
\includegraphics[width=\columnwidth]{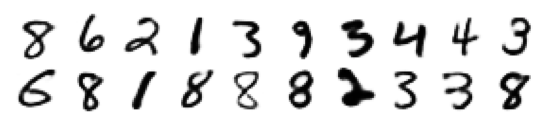}
\vspace{-6mm}
\caption{Generated samples for MNIST. 
For a video of the generation process, see {\tiny\url{https://youtu.be/ptLdYd8FXRA}}
}
\vspace{-5mm}
\label{fig:mnist-samples}
\end{figure}

We highlight the behaviour of the models using two data sets based on the MNIST benchmark.
The first experiment uses the binarized MNIST data set of \citet{salakhutdinov2008quantitative}, that consists of $28\times28$ binary images with 50,000 training and 10,000 test images. Table \ref{tab:mnist} compares the log-likelihoods on this binarized MNIST data set using existing models, as well as the models developed in this paper (with variances of our estimates in parentheses). The sequential generative model that uses the spatially-transformed attention with the CGRU hidden canvas provides the best performance among existing work on this data set. We show samples from the model in figure \ref{fig:mnist-samples}.

We form a multi-MNIST data set of $64 \times 64$ images that consists of two MNIST digits placed at random locations in the image (having adapted the cluttered MNIST generator from \citet{mnih2014recurrent} to procedurally generate the data). We compare the performance  in table \ref{tab:multimnist} and show samples from this model in figure \ref{fig:mnist-samples}. This data set is much harder than MNIST to learn, with much slower convergence. The additive canvas with spatially-transformed attention provides a reliable way to learn from this data.

\begin{table}[]
\centering
\caption{Test set negative log-likelihood on MNIST.}
\label{tab:mnist}
\begin{tabular}{llcrr}
\hline
\multicolumn{4}{c}{\textbf{Model}} & \textbf{Test NLL} \\ 
\hline
\multicolumn{5}{c}{\emph{\scriptsize{From \citet{gregor2015draw} and \citet{burda2015importance}}}}\\
\multicolumn{4}{l}{DBM 2hl } &  $\approx$84.62\\
\multicolumn{4}{l}{DBN 2hl} &  $\approx$84.55\\
\multicolumn{4}{l}{NADE} & 88.33  \\
%\multicolumn{4}{l}{EoNADE 2hl (128 orderings)} & 85.10 \\ 
%\multicolumn{4}{l}{EoNADE-5 2hl (128 orderings)} & 84.68 \\ 
\multicolumn{4}{l}{DLGM-VAE} & $\approx$ 86.60 \\ 
%\multicolumn{4}{l}{VAE + HVI 8 leapfrog steps} & $\approx$ 85.51 \\ 
%\multicolumn{4}{l}{DLGM/VAE + Norm Flow} & $\approx$ 85.10 \\ 
\multicolumn{4}{l}{VAE + HVI/Norm Flow} & $\approx$ 85.10 \\ 
\multicolumn{4}{l}{DARN} & $\approx$ 84.13 \\ 
%\multicolumn{4}{l}{MADE 2hl (32 masks)} & 86.64 \\
\multicolumn{4}{l}{DRAW (64 steps, no attention)} & $\leq$ 87.40 \\ 
\multicolumn{4}{l}{DRAW (64 steps, Gaussian attention)} & $\leq$ 80.97 \\ 
%\multicolumn{4}{l}{IWAE (2 layers; 1 particle )} & $\approx  85.33$ \\
\multicolumn{4}{l}{IWAE (2 layers; 50 particles )} & $\approx 82.90$ \\
\hline
\multicolumn{5}{c}{\emph{\scriptsize{Sequential generative models}}}\\
\textbf{Attention} & \textbf{Canvas} & \textbf{Steps} & \textbf{Train} & \textbf{Test NLL}   \\ 
\hline
Spatial tr. & CGRU & 80 &  78.5& $\leq$\textbf{80.5(0.3)} \\
Spatial tr. & Additive & 80 &  80.1& $\leq$81.6(0.4) \\
Spatial tr. & CGRU & 30 &  80.1&$\leq$81.5(0.4) \\
Spatial tr. & Additive & 30 &  79.1& $\leq$82.6(0.5) \\
Fully conn. & CGRU & 80 &  80.0& $\leq$98.7(0.8)\\
\hline
\end{tabular}
\vspace{-2mm}
\end{table}

\begin{figure}[t]
\begin{minipage}[b]{0.45\linewidth}
\centering
\includegraphics[height=2.7cm]{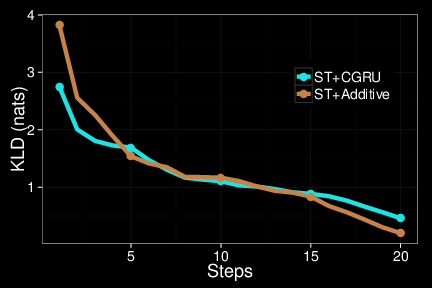}
%\vspace{-4mm}
\caption{Per-step KL contribution on MNIST.}
%\caption{Per-step KL $\KL[q_\phi(\vz_t | \vz_{1\ldots (t-1)} \vx) \| p(\vz_t)]$ contribution on MNIST.}
\label{fig:mnist-kld}
\end{minipage}
\hspace{2mm}
\begin{minipage}[b]{0.45\linewidth}
\centering
\includegraphics[height=2.7cm]{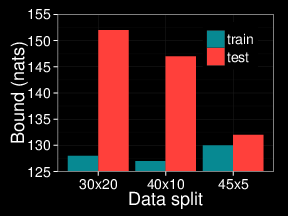}
%\vspace{-4mm}
\caption{Gap between train and test bound on omniglot.}
\label{fig:oneshot-overfitting}
\end{minipage}
\vspace{-4mm}
\end{figure}

%\vspace{1mm}
\textbf{Importance of each step}\\
These results also indicate that longer sequences can lead to better performance.
Every step taken by the model adds a term to the objective function \eqref{eq:FE} corresponding to the KL-divergence between the prior distribution and the contribution to the approximate posterior distribution at that step. Figure \ref{fig:mnist-kld} shows the KL-divergence for each iteration for two models on MNIST up to 20 steps. The KL-divergence decays towards the end of the sequence, indicating that the latent variables $\vz_t$ have diminishing contribution to the model as the number of steps grow. Unlike VAEs where we often find that there are many dimensions which contribute little to the likelihood bound, the sequential property allows us to more efficiently allocate and decide on the number of latent variables to use and means of deciding when to terminate the sequential computation.

%\begin{figure}[t]
%	\includegraphics[width=\columnwidth]{mnist-kld}
%	\caption{ KLD between prior and posterior at each iteration of a 20-step model, using additive and CGRU canvases.}
%	\label{fig:mnist-kld}
%\end{figure}

%\begin{figure}[t!]
% \centering
%\subfigure[Per-step KL contribution.]{
%\includegraphics[height=2.5cm]{mnist-kld}
%\label{fig:mnist-kld}
%}
%%\hspace{0.5cm}
%\subfigure[Gap between train and test bound.]{
%\includegraphics[height=2.5cm]{omniglot_samechar_overfitting}
%\label{fig:oneshot-overfitting}
%}
%\caption{REPLACE WITH MINIPAGE}
%\end{figure}

\vspace{-0.2cm}
\begin{table}[t]
	\begin{center}
		\caption{Train and test NLL bounds on $64\times 64$ Multi-MNIST.}
			\label{tab:multimnist}
		\begin{tabular}{lcccc}
			\hline
			\textbf{Att} & \textbf{CT} & \textbf{Steps} & \textbf{Train} & \textbf{Test}   \\ 
			\hline
			Multi-ST & Additive & 80 & \textbf{177.2}& \textbf{176.9(0.5)} \\
			Spatial tr. & Additive & 80 &  183.0& 182.0(0.6) \\
			Spatial tr. & CGRU & 80 &  196.0& 194.9(0.5) \\
			Fully conn. & CGRU & 80 &  272.0&270.3(0.8)\\
			\hline
		\end{tabular}
	\end{center}
\vspace{-7mm}
\end{table}

\begin{figure}[t]
\centering
\includegraphics[width=1\columnwidth]{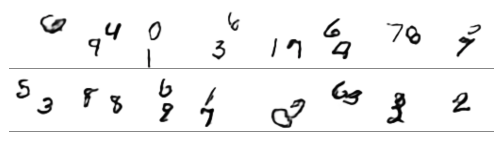}
\vspace{-4mm}
\caption{Generated samples for multi-MNIST. For a video of the generation process, see {\tiny\url{https://www.youtube.com/watch?v=HkDxmnIfWIM}}}
\vspace{-4mm}
\label{fig:multimnist-samples}
\end{figure}

\subsection{Omniglot}
Unlike MNIST, which has a small number of classes with many images of each class and a large amount of data, the omniglot data set \citep{lake2015human} consists of $105\times105$ binary images across 1628 classes with just 20 images per class. This data set allows us to demonstrate that attentional mechanisms and better generative models allow us to perform well even in regimes with larger images and limited amounts of data. 

There are two versions of the omniglot data that have been previously used for the evaluation of generative models. One data set used by \citet{burda2015importance} consists of $28\times28$ images, but is different to that of \citet{lake2015human}. We compare the available methods on the dataset from \citet{burda2015importance} in table \ref{tab:omniglot28} and find that the sequential models perform better than all competing approaches, further establishing the effectiveness of these models. Our second evaluation uses the dataset of \citet{lake2015human}, which we downsampled to $52\times52$ using a $2\times2$ max-pooling. We compare different sequential models in table \ref{tab:omniglot} and again find that spatially-transformed attention is a powerful general purpose attention and that the additive hidden canvas performs best. %Using an attention based on a fully-connected transformation is a poor choice. 

\vspace{-0.2cm}
\begin{table}[t]
\centering
\caption{NLL on the $28\times28$ omniglot data.}
\label{tab:omniglot28}
\begin{tabular}{lr}
\hline
\textbf{Model}& \textbf{Test NLL} \\
\hline
\multicolumn{2}{c}{\emph{\scriptsize{From \citet{burda2015importance}}}}\\
VAE (2 layer, 5 samples) & 106.31 \\
IWAE (2 layer, 50 samples) & 103.38 \\
RBM (500 hidden) & 100.46 \\
\hline
Seq Gen Model (20 steps, ST, additive) & $\leq$\textbf{96.5}\\
Seq Gen Model (80 steps, ST, additive) & $\leq$\textbf{95.5}\\
\hline
\vspace{-9mm}
\end{tabular}
\end{table}
\begin{table}[t]
	\begin{center}
		\caption{Train and test NLL bounds on $52\times 52$ omniglot}
			\label{tab:omniglot}
		\begin{tabular}{lcccc}
			\hline
			\textbf{Att} & \textbf{CT} & \textbf{Steps} & \textbf{Train} & \textbf{Test}   \\ 
			\hline
			Multi-ST & CGRU & 80 & \textbf{120.6}& \textbf{134.1(0.5)} \\
			Spatial tr.  & Additive & 40 &  128.7&136.1(0.4)  \\
			Spatial tr.  & Additive & 80 &  134.6&141.5(0.5) \\
			Spatial tr.  & CGRU & 80 &  141.6& 144.5(0.4) \\
			Fully conn. & CGRU & 80 &  170.0& 351.5(1.2)\\
			\hline
		\end{tabular}
	\end{center}
	\vspace{-5mm}
\end{table}

\subsection{Multi-PIE}
\begin{figure}[t]
\centering
	\includegraphics[width=0.6\columnwidth]{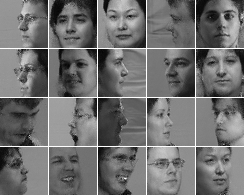}
	\vspace{-2mm}
	\caption{Generated samples for Multi-PIE using the model with Spatial Transformer + additive canvas (32 steps). For a video of the generation process including the boundaries of the writing attention grid, see  {\tiny\url{https://www.youtube.com/watch?v=6S6Tx_OtvnA}}
	}
	\vspace{-4mm}
	\label{fig:multipie-samples}
\end{figure}

The Multi-PIE dataset \cite{gross2010multi} consists of $48\times48$ RGB face images from various viewpoints. We have converted the images to grayscale and trained our model on a subset comprising of all 15-viewpoints but only 3 out of the 19 illumination conditions. Our simplification results in $93,130$ training samples and $10,000$ test samples. Samples from this model are shown in figure \ref{fig:multipie-samples} and are highly compelling, showing faces in different orientations, different genders and are representative of the data.
The model was trained using the logit-normal likelihood as in \citet{rezende2015variational}.

%\begin{table}[t]
%	\begin{center}
%	\caption{Test set negative log-likelihood on MNIST.}
%	\begin{tabular}{lr}
%		\hline
%		\textbf{Model} & \textbf{NLL} (\textit{nats})  \\ 
%		\hline
%		\multicolumn{2}{c}{\emph{\scriptsize{From \citet{gregor2015draw} and \citet{burda2015importance}}}}\\
%		DBM 2hl & $\approx$ 84.62 \\ 
%		DBN 2hl & $\approx$ 84.55 \\ 
%		NADE & 88.33 \\ 
%		EoNADE 2hl (128 orderings) & 85.10 \\ 
%		EoNADE-5 2hl (128 orderings) \quad \quad \quad & 84.68 \\ 
%		VAE & $\approx$ 86.60 \\ 
%		VAE + HVI 8 leapfrog steps & $\approx$ 85.51 \\ 
%		DLGM/VAE + Norm Flow & $\approx$ 85.10 \\ 
%		DARN 1hl & $\approx$ 84.13 \\ 
%		MADE 2hl (32 masks) & 86.64 \\
%		DRAW (40 steps) & $\leq$ 80.97 \\ 
%		\hline
%		\multicolumn{2}{c}{\emph{\scriptsize{From \citet{burda2015importance}}}}\\
%		IWAE (2 layers; 1 particles ) & $\approx  85.33$ \\
%		IWAE (2 layers; 50 particles ) & $\approx 82.90$ \\
%	    \hline
%	\end{tabular}
%	\end{center}
%\label{tab:mnist}
%\end{table}
%
%\begin{table}[t]
%	\begin{center}
%		\caption{Train and test NLL bounds on $28\times 28$ MNIST}
%		\begin{tabular}{lcccc}
%			\hline
%			\textbf{Att} & \textbf{CT} & \textbf{Steps} & \textbf{Train} & \textbf{Test}   \\ 
%			\hline
%			ST & CGRU & 80 &  \textbf{78.5}& \textbf{80.1} \\
%			ST & Additive & 80 &  \textbf{80.1}& \textbf{80.6} \\
%			ST & CGRU & 30 &  \textbf{80.1}&\textbf{81.0} \\
%			ST & Additive & 30 &  \textbf{79.1}& \textbf{81.6} \\
%			FC & CGRU & 80 &  \textbf{80.0}& \textbf{97.8}\\
%			\hline
%		\end{tabular}
%	\end{center}
%	\label{tab:mnist2}
%\end{table}

%% file: oneshot.tex
\vspace{-3mm}
\section{One-Shot Generalization}
\vspace{-0mm}
\citet{lake2015human} introduce three tasks to evaluate one-shot generalization, testing weaker to stronger forms of generalization. The three tasks are: (1) unconditional (free) generation, (2) generation of novel variations of a given exemplar, and (3) generation of representative samples from a novel alphabet. \citet{lake2015human} conduct human evaluations as part of their assessment, which is important in contrasting the performance of models against the cognitive ability of humans; we do not conduct human benchmarks in this paper (human evaluation will form part of our follow-up work). Our focus is on the machine learning of one-shot generalization and the computational challenges associated with this task.
%Our intended aim is to demonstrate a new class of models that possess the important ability for one-shot generalization, which has thus far been demonstrated by very few other machine learning approaches. We obtain this ability by constructing powerful sequential generative models that incorporate the principles of feedback and attention established in the previous section. Of the five tasks developed by \citet{lake2015human}, three are related to the problem of one-shot inference and generation with the omniglot dataset, and we demonstrate the ability of this model on all three of these tasks. The three tasks are: unconditional (free) generation, generation of novel variations of a given exemplar, and generation of representative samples from a novel alphabet. 

\begin{figure}[t]
\centering
\includegraphics[width=0.85\columnwidth]{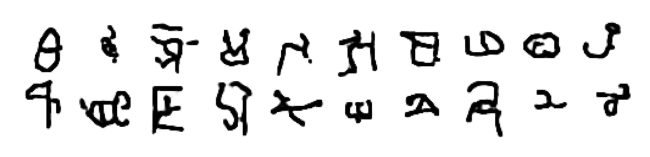}
\vspace{-6mm}
\caption{Unconditional samples for $52\times52$ omniglot (task 1). 
For a video of the generation process, see {\tiny\url{https://www.youtube.com/watch?v=HQEI2xfTgm4}}
}
\vspace{-2mm}
\label{fig:omniglot-samples}
\end{figure}

\textbf{1. Unconditional Generation.}\\
This is the same generation task reported for the data sets in the previous section. Figure \ref{fig:omniglot-samples} shows samples that reflect the characteristics of the omniglot data, showing a variety of styles including rounded patterns, line segments, thick and thin strokes that are representative of the data set. The likelihoods reported in tables \ref{tab:omniglot28} and \ref{tab:omniglot} quantitatively establish this model as among the state-of-the-art. % in this task.

\begin{figure}[t]
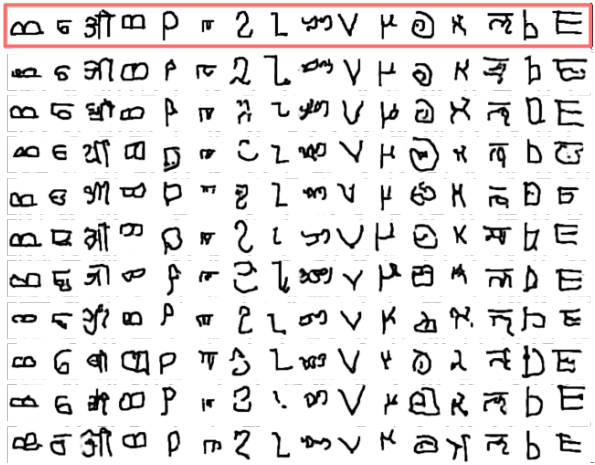

\centering
%\adjincludegraphics[height = 4.5cm, trim={0 0 {0.335\width} 0},clip]{negated/oneshot-samples-samechar-icml}
\adjincludegraphics[width=0.95\columnwidth, trim={0 0 {0\width} 0},clip]{negated/oneshot-samples-samechar-icml.png}

\vspace{-2mm}
\caption{Generating new examplars of a given character for the weak generalization test (task 2\textit{a}).
The first row shows the test images and the next 10 are one-shot samples from the model. 
}
\vspace{-6mm}
\label{fig:oneshot-generation-weak}
\end{figure}

\begin{figure}[t]
\centering

%\adjincludegraphics[width=1\columnwidth, trim={0 0 {0.\width} 0},clip]{negated/oneshot-samples-samechar-multidataset}
%\vspace{-5mm}
\begin{overpic}[width=\columnwidth,tics=10]{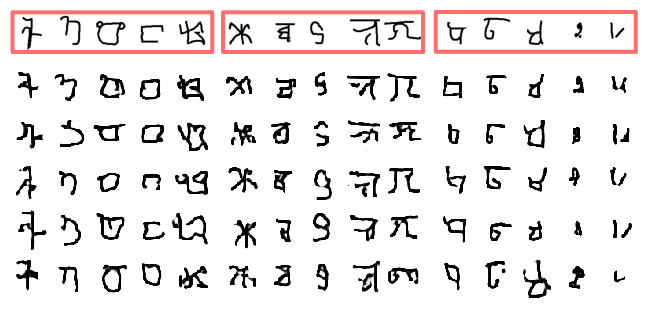}
\put(15, 48 ) {\tiny \textbf{30-20}}
\put(48, 48 ) {\tiny \textbf{40-10}}
\put(80, 48 ) {\tiny \textbf{45-5}}
\end{overpic}
\vspace{-5mm}
\caption{Generating new examplars of a given character for the strong generalization test (task 2\textit{b,c)}, with models trained with different amounts of data. Left: Samples from model trained on 30-20 train-test split; Middle: 40-10 split; Right: 45-5 split (right)}
\label{fig:oneshot-generation-strong}
\vspace{-2mm}
\end{figure}

\textbf{2. Novel variations of a given exemplar.}\\
This task corresponds to figure 5 in \citet{lake2015human}). At test time, the model is presented with a character of a type it has never seen before (was not part of its training set), and asked to \textit{generate novel variations of this character}. 
To do this, we use a conditional generative model (figure \ref{fig:condModel}, equation \eqref{eq:mod_context}). The context $\vx'$ is the image that we wish the model to generate new exemplars of.  To expose the boundaries of our approach, we test this under weak and strong one-shot generalization tests:
\begin{enumerate}[leftmargin=*, label=\emph{\alph*)}]
\item We use a data set whose training data consists of all available alphabets, but for which three character types from each alphabet have been removed to form the test set (3000 characters). This is a weak one-shot generalization test where, although the model has never seen the test set characters, it has seen related characters from the same alphabet and is expected to transfer that knowledge to this generation task.
\item We use exactly the data split used by \citet{lake2015human}, which consists of 30 alphabets as the training set and the remaining 20 alphabets as the test set. This is a strong one-shot generalization test, since the model has seen neither the test character nor any alphabets from its family. This is a hard test for our model, since this split provides limited training data, making overfitting easier, and generalization harder.
\item We use two alternative training-test split of the data, a 40-10 and 45-5 split. We can examine the spectrum of difficulty of the previous one-shot generalization task by considering these alternative splits.
\end{enumerate}
We show the model's performance on the weak generalization test in figure \ref{fig:oneshot-generation-weak}, where the first row shows the exemplar image, and the subsequent rows show the variations of that image generated by the model. We show generations for the strong generalization test in figure \ref{fig:oneshot-generation-strong}. Our model also generates visually similar and reasonable variations of the image in this case. Unlike the model of \citet{lake2015human}, which uses human stroke information and a model structured around the way in which humans draw images, our model is applicable to any image data, with the only domain specific information that is used being that the data is spatially arranged (which is exploited by the convolution and attention). This test also exposes the difficulty that the model has in coping with small amounts of data. We compare the difference between train and test log-likelihoods for the various data splits in figure \ref{fig:oneshot-overfitting}. We see that there is a small gap between the training and test likelihoods in the regime where we have more data (45-5 split) indicating no overfitting. There is a large gap for the other splits, hence a greater tendency for overfitting in the low data regime. 
An interesting observation is that even for the cases where there is a large gap between train and test likelihood bounds (figure 5), the examples generated by the model (figure 10, left and middle) still generalize to unseen character classes.
Data-efficiency is an important challenge for the large parametric models that we use and one we hope to address in future.

\begin{figure}[t]
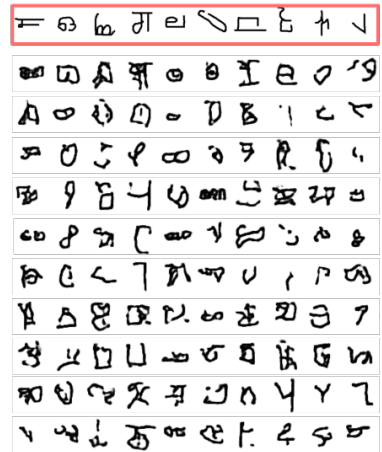

\centering
\adjincludegraphics[width=0.65\columnwidth, trim={0 0 {0.55\width} 0},clip]{negated/oneshot-samples-samealphabet-30x20x1}
\vspace{-2mm}
\caption{Generating new exemplars from a novel alphabet (task 3). The first row shows the test images, and the next 10 rows are one-shot samples generated by the model. }
\vspace{-4mm}
\label{fig:oneshot-newalpabet}
\end{figure}

\vspace{5mm}
\textbf{3. Representative samples from a novel alphabet.}\\
This task corresponds to figure 7 in \citet{lake2015human}, and conditions the model on anywhere between 1 to 10 samples of a novel alphabet and asks the model to \textit{generate new characters consistent with this novel alphabet}. We show here the hardest form of this test, using only 1 context image. This test is highly subjective, but the model generations in figure \ref{fig:oneshot-newalpabet} show that it is able to pick up common features and use them in the generations. 

We have emphasized the usefulness of deep generative models as  scalable, general-purpose tools for probabilistic reasoning that have the important property of one-shot generalization. But, these models do have limitations. We have already pointed to the need for reasonable amounts of data. Another important consideration is that, while our models can perform one-shot generalization, they do not perform one-shot \textit{learning}. One-shot learning requires that a model is updated after the presentation of each new input, e.g., like the non-parametric models used by \citet{lake2015human} or \citet{salakhutdinov2013learning}. Parametric models such as ours require a gradient update of the parameters, which we do not do. Instead, our model performs a type of one-shot inference that during test time can perform inferential tasks on new data points, such as missing data completion, new exemplar generation, or analogical sampling, but does not learn from these points. This distinction between one-shot learning and inference is important and affects how such models can be used. We aim to extend our approach to the online and one-shot learning setting in future.

%% file: discussion.tex
\vspace{-2mm}
\section{Conclusion}
\vspace{-0mm}
 %There are a number of direct extension that this setting can be combined with. 
%\begin{itemize}
%\item Use of this framework in multi-sample loss functions.
%\end{itemize}
We have developed a new class of general-purpose models that have the ability to perform one-shot generalization, emulating an important characteristic of human cognition. Sequential generative models are natural extensions of variational auto-encoders and provide state-of-the-art models for  deep density estimation and image generation. The models specify a sequential process over groups of latent variables that allows it to compute the probability of data points over a number of steps, using the principles of feedback and attention. 
The use of spatial attention mechanisms substantially improves the ability of the model to generalize.
The spatial transformer is a highly flexible attention mechanism for both reading and writing, and is now our default mechanism for attention in generative models. 
%We demonstrated that the models have very good performance over a wide range of data sets, consisting of images of various size and color channels, as well as with varying amounts of data, showing the ability to learn under very different conditions. 
We highlighted the one-shot generalization ability of the model over a range of tasks that showed that the model is able to generate compelling and diverse samples, having seen new examples just once.  However there are limitations of this approach, e.g., still needing a reasonable amount of data to avoid overfitting, which we hope to address in future work.

%% file: acknowledgements.tex
\section*{Acknowledgements}
We thank Brenden Lake and Josh Tenenbaum for insightful discussions. We are grateful to Theophane Weber, Ali Eslami, Peter Battaglia and David Barrett for their valuable feedback.

%% file: appdxSamples.tex
\section{Additional Results}
\label{appdx:addResults}

\subsection{SVHN}
The SVHN dataset \cite{netzer2011reading} consists of $32\times32$ RGB images from house numbers. 

\begin{figure}[t]
	\includegraphics[width=\columnwidth]{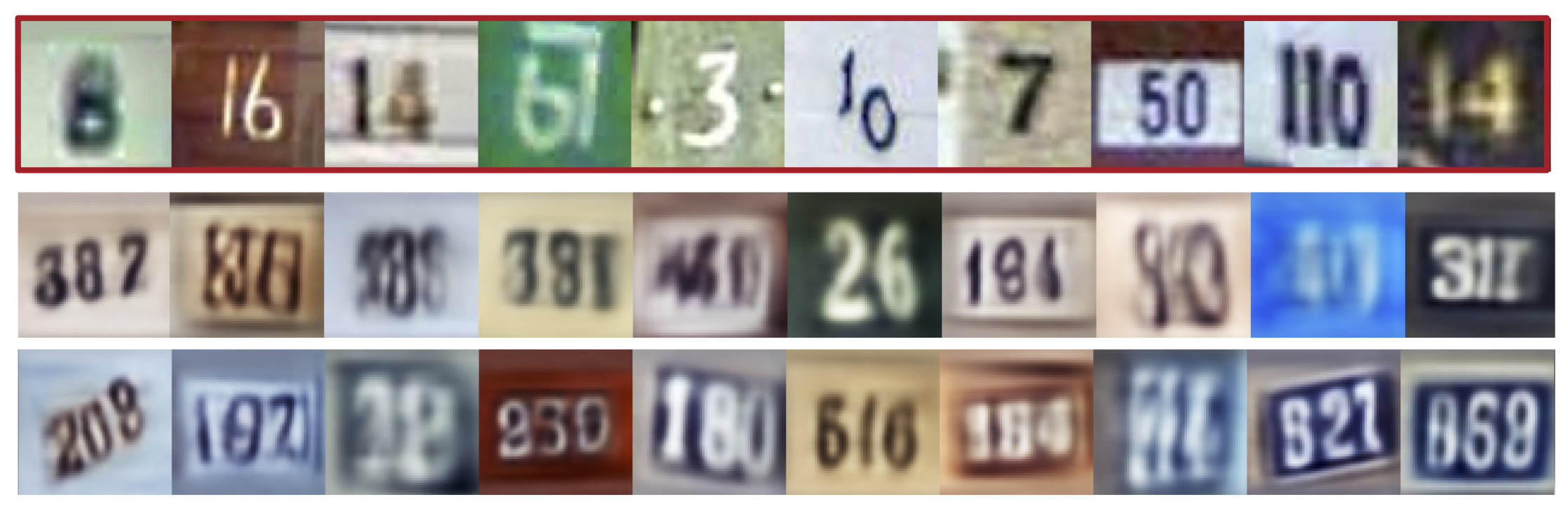}
	\caption{Generated samples for SVHN using the model with Spatial Transformer + Identity (80 steps). 
	For a video of the generation process, see this video {\tiny \url{https://www.youtube.com/watch?v=281wqqkmAuw}}
	}
	\label{fig:multimnist-samples}
\end{figure}

%% file: appdxAttention.tex
\section{Other types of attention}
\label{appdx:attention}
\textbf{\textit{Randomized attention.}} The simplest attention randomly selects patches from the input image, which is the simplest way of implementing a sparse selection mechanism. Applying dropout regularisation to the input layer of deep models would effectively implement this type of attention (a hard attention that has no learning). In data sets like MNIST this attention allows for competitive learning of the generative model if the model is allowed to attend to a large number of patches; see this video {\tiny \url{https://www.youtube.com/watch?v=W0R394wEUqQ}}.

\textbf{\textit{Error-based attention}}. One of the difficulties with attention mechanisms is that for large and sparse images, there can be little gradient information available, which can cause the attentional selection to become stuck. To address this issue, previous approaches have used particle methods \cite{tang2014learning} and exploration techniques from reinforcement learning \cite{mnih2014recurrent} to infer the latent variables that control the attentional, and allow it to jump more easily to relevant parts of the input. A simple way of realizing this, is to decide where to attend to by jumping o places where the model has made the largest reconstruction errors.  To do this, we convert the element-wise reconstruction error at every step into a probability map of locations to attend to at the next iteration:
$$p( a_t=k | \vx, \hat{\vx}_{t-1} ) \propto \exp\left(-\beta | \frac{\epsilon_k - \bar{\epsilon}}{ \kappa } | \right)$$
where $\epsilon_k = x_k - \hat{x}_{t-1, k}$ is the reconstruction error of the $k$th pixel, $\hat{\vx}_{t-1}$ is the reconstructed image at iteration $t-1$, $\vx$ is the current target image, $\bar{\epsilon}$ is the spatial average of $\epsilon_k$, and $\kappa $ is the spatial standard deviation of $\epsilon_k$. This attention is %highly 
suited to models of sparse images 
%The images should not be too larges, since computing the attention probabilities will scale with the size of the images, which will become prohibitive when the images are very large
; see this video {\tiny \url{https://www.youtube.com/watch?v=qb2-73OHuWA}} for an example of a model with this attention mechanism.
In this type of hard-attention, a policy does not need to be learned, since a new one is obtained after every step based on the reconstruction error and effectively allows every step to work more efficiently towards reducing the reconstruction error. It also overcomes the problem of limited gradient information in large, sparse images, since this form of attention will have a saccadic behaviour since it will be able to jump to any part of the image that has high error.

\textbf{\textit{Multiple spatial attention}}.
A simple generalization of using a single spatial transformer is to have multiple STs that are additively combined:
\begin{align}
	\vy(\vv) &= \sum_{i=1}^K\left[ \kappa_h(\vh_i(\vv)) \otimes \kappa_w(\vh_i(\vv)) \right]   \ast  \vx_i(\vv), \nonumber
\end{align}
where $\vv$ is a context that conditions all STs. This module allows the generative model to write or read at multiple locations simultaneously. 
%The next section will develop a new class of generative models and these mechanisms will be incorporated into this framework providing many different generative models that will have advantages in different modelling scenarios.

\section{Other model details}

The CGRU of \citet{kaiser2015neural} has the following form:
\begin{align}
& f_c(\vc_{t-1}, \vh_t;\theta_c) = \text{CGRU}( \vc_{t-1} +  f_w(  \vh_t;\theta_c ) ), \label{eq_cgru_canvas} \\
& \text{CGRU}(c) = \vu \odot \vc + (1 - \vu)\odot \tanh( \vU \ast  ( \vr \odot \vc) + \vB  ), \nonumber\\
& \vu =   \sigma( \vU' \ast \vc + \vB' ), \quad \vr =  \sigma( \vU'' \ast \vc + \vB'' ) \nonumber
\end{align}
where the symbols $\odot$ indicates the element-wise product, $\ast$ a size-preserving convolution with stride of $1\times1$, and $\sigma(\cdot)$ is the sigmoid function. The matrices $U$, $U'$ and $U''$ are $3\times3$ kernels. The number of filters used for the hidden canvas $\vc$ is specified on section \ref{sec.results}.